\documentclass[letterpaper, 10 pt, conference]{ieeeconf} 
\IEEEoverridecommandlockouts
\usepackage{cite}
\usepackage{amsmath,amssymb,amsfonts}
\usepackage{bm} 
\usepackage{subfigure} 
\usepackage{enumerate}
\usepackage{url}
\usepackage{algorithmic}
\usepackage{graphicx}
\usepackage{textcomp}
\usepackage{xcolor}
\def\BibTeX{{\rm B\kern-.05em{\sc i\kern-.025em b}\kern-.08em
    T\kern-.1667em\lower.7ex\hbox{E}\kern-.125emX}}
    
\begin{document}

\title{\LARGE \bf
CoBigICP: Robust and Precise Point Set Registration using Correntropy Metrics and Bidirectional Correspondence}
\author{Pengyu Yin$^{\dagger 1}$, Di Wang$^{\dagger 1}$, Shaoyi Du$^{*1}$, Shihui Ying$^2$, Yue Gao$^{*3}$, and Nanning Zheng$^1$, \emph{IEEE Fellow}  
\thanks{This work was supported by the National Key Research and Development Program of China under Grant No. 2017YFA0700800, the National Natural Science Foundation of China under Grant Nos. 61971343, 61790563 and 61627811, the Natural Science Basic Research Plan in Shaanxi Province of China under Grant No. 2020JM-012.}
\thanks{$^\dagger$ The authors contribute equally to this paper.}
\thanks{$^1$ Authors are with Institute of Artificial Intelligence and Robotics, College of Artificial Intelligence, Xi'an Jiaotong University, Shaanxi 710049, P.R. China. }
\thanks{$^2$ S. Ying is with Shanghai University, Shanghai 200444, P.R. China. }
\thanks{$^3$ Y. Gao is with Tsinghua University, Beijing 10048, P.R. China. }
\thanks{*Emails: {\tt\small \{dushaoyi,kevin.gaoy\}@gmail.com}}}

\maketitle

\begin{abstract}
In this paper, we propose a novel probabilistic variant of iterative closest point (ICP) dubbed as CoBigICP. The method leverages both local geometrical information and global noise characteristics. Locally, the 3D structure of both target and source clouds are incorporated into the objective function through bidirectional correspondence. Globally, error metric of correntropy is introduced as noise model to resist outliers. Importantly, the close resemblance between normal-distributions transform (NDT) and correntropy is revealed. To ease the minimization step, an on-manifold parameterization of the special Euclidean group is  proposed. Extensive experiments validate that CoBigICP outperforms several well-known and state-of-the-art methods. 
\end{abstract}


\section{Introduction}
Point cloud registration aims at finding the rigid transformation between two point clouds. It is an essential technique in many fields including reconstruction, medical application, and localization for mobile robots. One of the most famous and effective methods to solve the point cloud registration problem is iterative closest point (ICP) \cite{10.1007/s10514-013-9327-2}. 
On the whole, a general workflow of ICP consisting of 1) data point filtering, 2) data association, 3) outlier filtering and 4) error minimization. Many researchers focus on the third stage to realize a robust and precise algorithm: as a large enough portion of outlier could have more impact on the error minimization step than the inlier, results in a misalignment in the final result.

The most common outlier filtering-based algorithms are Generalized-ICP (GICP) \cite{inproceedings} and its variants. Instead of using single points for registration, these approaches embed the geometric structure into the objective function. GICP emphasizes the residual in the normal direction while neglects the co-planar residual. A well-known GICP variant is normal ICP (NICP) \cite{7353455}, who additionally incorporated the normal vector into the objective function with a simplification to the information matrix. Tabib et al. \cite{8411140} proposed a Gaussian mixture model (GMM) based registration technique that forms a more precise weight distribution for point-pair. Parkison et al. \cite{parkison2018semantic} converted the solution space from Euler angle to special Euclidean and adopted Cauchy loss as the error function. Wang et al. \cite{8594278} proposed a mixture of exponential power distributions to model the residual error. However, these methods lack attention to the correspondence establishment and suffer from heavy outliers. Other solutions exist that rely on thresholds \cite{1047997} or continuous functions.

\begin{figure}
\centering
\subfigure[GT+CoBigICP]{\includegraphics[height=2.6cm]{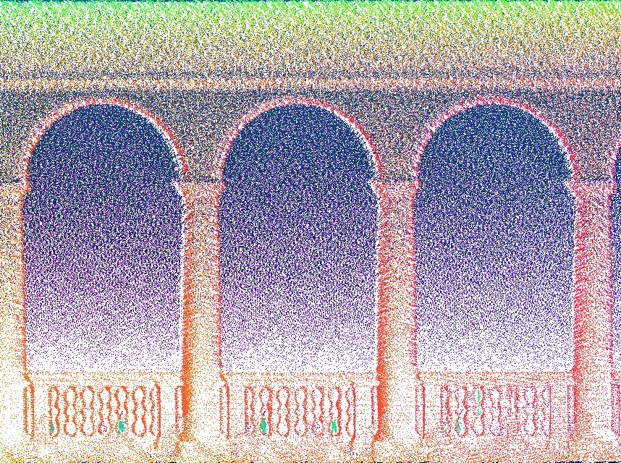}}
\hspace{0.1in}
\subfigure[GT+GICP]{\includegraphics[height=2.6cm]{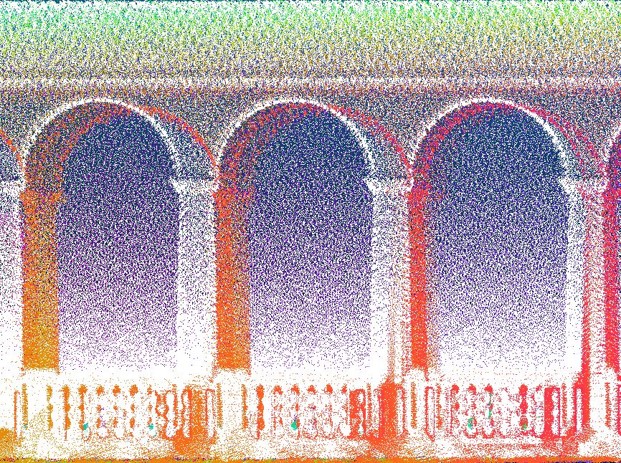}}
\caption{ Reconstruction results of ETH Hauptgebaude with CoBigICP and GCIP are presented. To be more illustrative, with the premise of the point clouds are put in the same coordinate, we lay the reconstruction point cloud of CoBigICP and GICP (both are colored) on the GT point cloud (white points) separately with no post-processing. We observe clearly the misalignment in the result of GICP and a nearly perfect reconstruction result with CoBigICP.}
\label{Fig recon res}
\vspace{-1.5em}
\end{figure}

In \cite{zaganidis2018integrating}, the similarity of GICP and NDT is noticed. The intuition of our work stems from the robust noise modeling that implicitly adopted in normal-distributions transform (NDT)\cite{1249285}, and the probabilistic model of GICP.

Contributions of this paper are as follows: 
\begin{enumerate}
    \item The proposed algorithm 
    leverages bidirectional correspondences for a symmetric model.
    \item Correntropy is employed as a robust error metric, with its efficacy being inferred and proved. We revealed the insight of the close resemblance between NDT and correntropy.
    \item An on-manifold parameterization solution is provided.
\end{enumerate}
The rest of the paper is organized as follows: firstly in section \uppercase\expandafter{\romannumeral2} we introduce a general framework of ICP and the basic algorithm. In section \uppercase\expandafter{\romannumeral3} we detail the formulation of the proposed method and also the relationship between correntropy and NDT. In section \uppercase\expandafter{\romannumeral4} we present comparative experiments on different data sets. And finally in section \uppercase\expandafter{\romannumeral5} we draw the conclusion and discuss future works. 

An comparative reconstruction result is shown in Fig.~(\ref{Fig recon res}) and the source code is available at \url{https://github.com/Pamphlett/CoBigICP}.

\section{Problem Formulation}

\subsection{Mathematical formulation of point set registration}
Given two point clouds $\mathcal{A}$ and $\mathcal{B}$, point cloud registration problem aims at finding a rigid transformation that best aligns the \emph{source} point cloud $\mathcal{B}=\{{\bf b}_{1:N_{s}}\}$ to the \emph{target} point cloud $\mathcal{A}=\{{\bf a}_{1:N_{t}}\}$ \cite{zaganidis2018integrating}, where $\bf a$ and $\bf b$ denote the 3D points. We define the association variable $\mathcal{I}\overset{\Delta}{=}\left\{m_i,n_i\right\}_{i=1}^{N}\in \mathcal{I}_g$ where $m_i$, $n_i$ indicate ${\bf a}_{i}\overset{\Delta}{=}{\bf a}_{m_i}\in \mathcal{A}$ is a measurement of the same point as ${\bf b}_{i}\overset{\Delta}{=}{\bf b}_{n_i}\in \mathcal{B}$, and $\mathcal{I}_g$ is the set of all possible pair-wise correspondences. A general framework of ICP \cite{121791} is below:
\begin{equation}
    \mathop{\min}_{\bf T} \sum\limits_{i=1}^{N}
    \left({\bf a}_{i}-\mathbf{T}{\bf b}_{i}\right)^\top\mathbf{\Omega}_{i}
    \left({\bf a}_{i}-\mathbf{T}{\bf b}_{i}\right),
    \label{eq:G_Framework}
\end{equation}
where ${\bf{T}}\in {\rm SE(3)}$ denotes a rigid transformation that maps $\mathcal{B}$ to the $\mathcal{A}$, namely, ${\bf T}{\bf a}_i = {\bf R}{\bf b}_i + {\bf t}$, $\mathbf{R}\in {\rm SO(3)}$, ${\bf{t}}\in\mathbb{R}^{3}$ are rotation matrix and translation vector. $\mathbf{\Omega}_{i}$ denotes the information matrix that captures the local geometrical shape of the point pair ${\bf a}_i$ and ${\bf b}_{i}$ and ${\bf e}_i$ denotes the pair-wise error ${\bf e}_i={\bf a}_{i}-\mathbf{T}{\bf b}_{i}$. The original version of ICP algorithm \cite{121791} defines $\mathbf{\Omega}_{i}=\bf{I}$. For the variant who applies point-to-plane metric, it is the case where $\mathbf{\Omega}_{i}={\bf n}_{i}{\bf n}_{i}^{\top}$, with ${\bf n}_{i} \in\mathbb{R}^3 $ the surface normal of point ${\bf a}_{i}$.

The ICP algorithm can be summarized into two steps:
\begin{enumerate}
    \item Correspondence step: is to find a putative correspondence between two scans.
    \item Minimization step: is to compute a transformation via minimization of differences between corresponding points.
\end{enumerate}
In practice, the exact correspondence in correspond step is usually unknown. But some approximations like nearest neighbor are proven to be effective under the premise of a good initial value. In the minimization step the difference between two point clouds are usually modeled by the sum of squared errors. 

From a probabilistic perspective, the objective function of ICP can be reformulated to a well-defined parameter estimation problem and can be efficiently solved via maximum likelihood estimation (MLE). To be more specific, each pair-wise distance error is view as an independent measurement. A proper probabilistic distribution is utilized to generate all these measurements (noise modeling).
Then minimizing the objective function (\ref{eq:G_Framework}) is equivalent to maximizing the log likelihood function. 


We remark here that GICP, NICP and NDT \cite{1249285} are within the aforementioned general ICP framework (\ref{eq:G_Framework}). With the noise modeling method of GICP and NDT being different.



\subsection{Drawbacks of Previous Methods}


GICP attaches a probabilistic model to the minimization step. More explicitly, sets of underlying points are modeled by assigning Gaussian distributions to each of the points, ${\bf a}_m \sim \mathcal{N}\left(\mathbf{\hat{a}}_m, \bm{\Sigma}_{m,A}\right)$, ${\bf b}_n \sim \mathcal{N}(\mathbf{\hat{b}}_n, \bm{\Sigma}_{n,B})$. $\bm{\Sigma}_{m,A}$ and $\bm{\Sigma}_{n,B}$ stand for covariance matrices of each centered point set. With a correspondence $\mathcal{I}$ computed by nearest neighbor, the pair-wise error can be derived as $\mathbf{e}_i \sim \mathcal{N}\left(\mathbf{0},\bm{\Sigma}_{i,A} + \mathbf{R}\bm{\Sigma}_{i,B}\mathbf{R}^\top\right)$. In this context, any pair-wise error is drawn from the above multi-variate Gaussian distribution. Therefore, with MLE, the negative log-likelihood function is derived:
\begin{equation}
    \mathbf{T}^{\ast}=\mathop{\arg\min}_\mathbf{T}\sum\limits_{i=1}^{N}
    \mathbf{e}_{i}^\top{\left(\mathbf{\Sigma}_{i,A}+\mathbf{R}\mathbf{\Sigma}_{i,B}\mathbf{R}^\top\right)^{-1}}\mathbf{e}_{i},
    \label{eq:GICP}
\end{equation}
where the smallest eigenvalue of the two covariance matrices are replaced by a small constant $\epsilon$ to force points to lie on a local planar. 

Like many ICP variants, GICP employs heuristic correspondence (nearest neighbor) which is not robust to outliers. Moreover, the information matrix in (\ref{eq:GICP}) is computationally expensive. Also the Gaussian assumption is brittle in complex scenarios where large portion of outliers pervasively exist.

As for NDT, the algorithm firstly subdivide the reference scan space into a grid of cells. For each cell, a PDF is computed to approximate the local surface.
Then the goal of NDT is to find the best pose transform $\mathbf{T^{\ast}}$ that maximize the likelihood of the the current scan points lie on the reference surface. 

Importantly, the PDF of the cells is considered to be a mixture of normal distribution and a uniform distribution (in \cite{magnusson2009three} Chapter 6) to take the distribution of outliers into consideration:
\begin{equation}
    \widetilde{p}({\bf x})=c_1\mathcal{N}(\bm{\mu},\Sigma)+
    c_2p_0,
\end{equation}
where $c_1$ and $c_2$ are constants and can be computed based on the user-specified outlier ratio.

NDT also neglects the importance of the correspondence step and adopts an implicit nearest neighbor. The spurious correspondence loses accuracy and the fixed portion of outlier makes NDT nonflexible.

\section{Proposed Algorithm}

Our method improves upon both steps of ICP framework. Fundamentally, a similar probabilistic model of GICP is drawn to every single point. For the correspondence step, we propose a bidirectional correspondence to ameliorate heuristic correspondence. For the second step, we adopt a robust noise model and derive correntropy as the error metric.

\subsection{Re-Attaining Symmetry via Bidirectional Correspondence}
We incorporate both point clouds local structure into the objective function in a two-way process, namely bidirectional correspondence. The bidirectional correspondence includes two parts: the bidirectional search and the bidirectional distance residual. Take ${\bf a}_i$ in the source point cloud and its corresponding point ${\bf b}_{i}$ in the target point cloud as an example. The bidirectional search is with the following mathematical formulation:
\begin{small}
\begin{align}
    \begin{split}
    \mathcal{I}_{bi}=\left\{m_i,n_i | m_i=i,n_i=\mathcal{C}^*_f(i),\left\|{\bf a}_{\mathcal{C}^*_{b}(i)}-{\bf a}_{i}\right\|_2<\epsilon\right\}_{i=1}^N,
    \end{split}
\end{align}
\end{small}
where $\mathcal{I}_{bi}$ is an association variable denoting a set of putative correspondences, $m_i,n_i$ are indices of the corresponding points and $\epsilon$ is an Euclidean distance bound. We adopt the following formulation to build the forward correspondence:
\begin{equation}
    \mathcal{C}^*_f(i)=\mathop{\arg\min}_{\mathcal{C}_f(i)\in
    \left\{1,2,3,\cdots N_{s}\right\}}
    \left(\left\|{\bf a}_{i}-\mathbf{T}{\bf b}_{\mathcal{C}_f(i)}\right\|_2^2\right),
\end{equation}
where $\left\|\cdot\right\|_2$ is the Euclidean norm. Conversely, we can compute the backward correspondence $\mathcal{C}^*_{b}(i)$ based on the acquisition of the forward correspondence as:
\begin{equation}
    \mathcal{C}_{b}^*(i)=\mathop{\arg\min}_{\mathcal{C}_{b}(i)\in
    \left\{1,2,3,\cdots N_{t}\right\}}
    \left(\left\|{\bf a}_{\mathcal{C}_{b}(i)}-\mathbf{T}{\bf b}_{\mathcal{C}^*_f(i)}\right\|_2^2\right).
\end{equation}
By taking this measure, quantity of involving points is reduced while correspondence is also ameliorated.

For the minimization step, to incorporate local information into the objective function, we propose the bidirectional distance residual. Firstly, we adopt the same approximation as NICP \cite{7353455}: neglect the covariance matrix of the source point cloud. This leads to a fixed information matrix during the whole registration process. We denote $\mathcal{I}_{bi}$ as the result of the bidirectional search and the uni-directional distance objective function can be formulated as: 
\begin{equation}
    \mathcal{J}_{uni}(i)=\left({\bf a}_{i}-\mathbf{T}{\bf b}_{i}\right)^\top\mathbf{\Omega}_{i,A}\left({\bf a}_{i}-\mathbf{T}{\bf b}_{i}\right).
\end{equation}

Inversely, to incorporate the structure of the target point cloud, the following bidirectional distance objective function is derived:
\begin{align}
    \mathcal{J}_{bi}(i)&=\left({\bf a}_{i}-\mathbf{T}{\bf b}_{i}\right)^\top\mathbf{\Omega}_{i,A}\left({\bf a}_{i}-\mathbf{T}{\bf b}_{i}\right) \\
    &+\left({\bf b}_{i}-\mathbf{T}^{-1}{\bf a}_{i}\right)^\top\mathbf{\Omega}_{i,B}\left({\bf b}_{i}-\mathbf{T}^{-1}{\bf a}_{i}\right).
\end{align}
With ${\bf b}_{i}-\mathbf{T}^{-1}{\bf a}_{i}=\mathbf{T}^{-1}\left(\mathbf{T} {\bf b}_{i}-{\bf a}_{i}\right)$, the above equation can be simplified as:
\begin{equation}
    \mathcal{J}_{bi}(i)\!=\!\left({\bf a}_{i}\!-\!\mathbf{T}{\bf b}_{i}\right)^\top\left(\mathbf{\Omega}_{i,A}\!+\!{\bf R}\mathbf{\Omega}_{i,B}{\bf R}^\top\right)\left({\bf a}_{i}\!-\!\mathbf{T}{\bf b}_{i}\right),
\end{equation}
with
\begin{equation}
    \mathbf{\Omega}_{i}=\mathbf{\Omega}_{i,A}+{\bf{R}}\mathbf{\Omega}_{i,B}{\bf{R}}^\top
    \label{propo Informat}
\end{equation}
being the bidirectional information matrix. Recall the information matrix of GICP is:
\begin{equation}
    \mathbf{\Omega}_{i}^G=\left(\mathbf{\Omega}_{i,A}^{-1}+{\bf R}\mathbf{\Omega}_{i,B}^{-1}{\bf R}^\top\right)^{-1}.
    \label{GICP InforMat}
\end{equation}
More explicitly, the bidirectional residual is achieved as follows: by fixing the target point cloud and applying the transformation to the source point cloud, we get the forward residual error and by fixing the source point cloud and applying the inverse transformation to the target point cloud, we get the backward residual error.
Compare (\ref{GICP InforMat}) with (\ref{propo Informat}), the latter can be view as an approximation of the original GICP one. We avoid to compute one time of matrix inversion with the aforementioned approximation and ease the computation of gradient.

The intuition of the above formation comes from a neutrality thinking. Given a correspondence between a point-pair, it is not reasonable to emphasize any of them. The source point cloud contains equivalent amount of information with respect to the target point cloud. Both of the two point clouds should be treated equally and have an equal contribution to the final objective function. Consequently, information of both point clouds have been considered and a bidirectional correspondence have been established.

\subsection{Correntropy-Based Noise Modelling}
\begin{figure}
\centering
\subfigure[MSE]{\includegraphics[height=3cm]{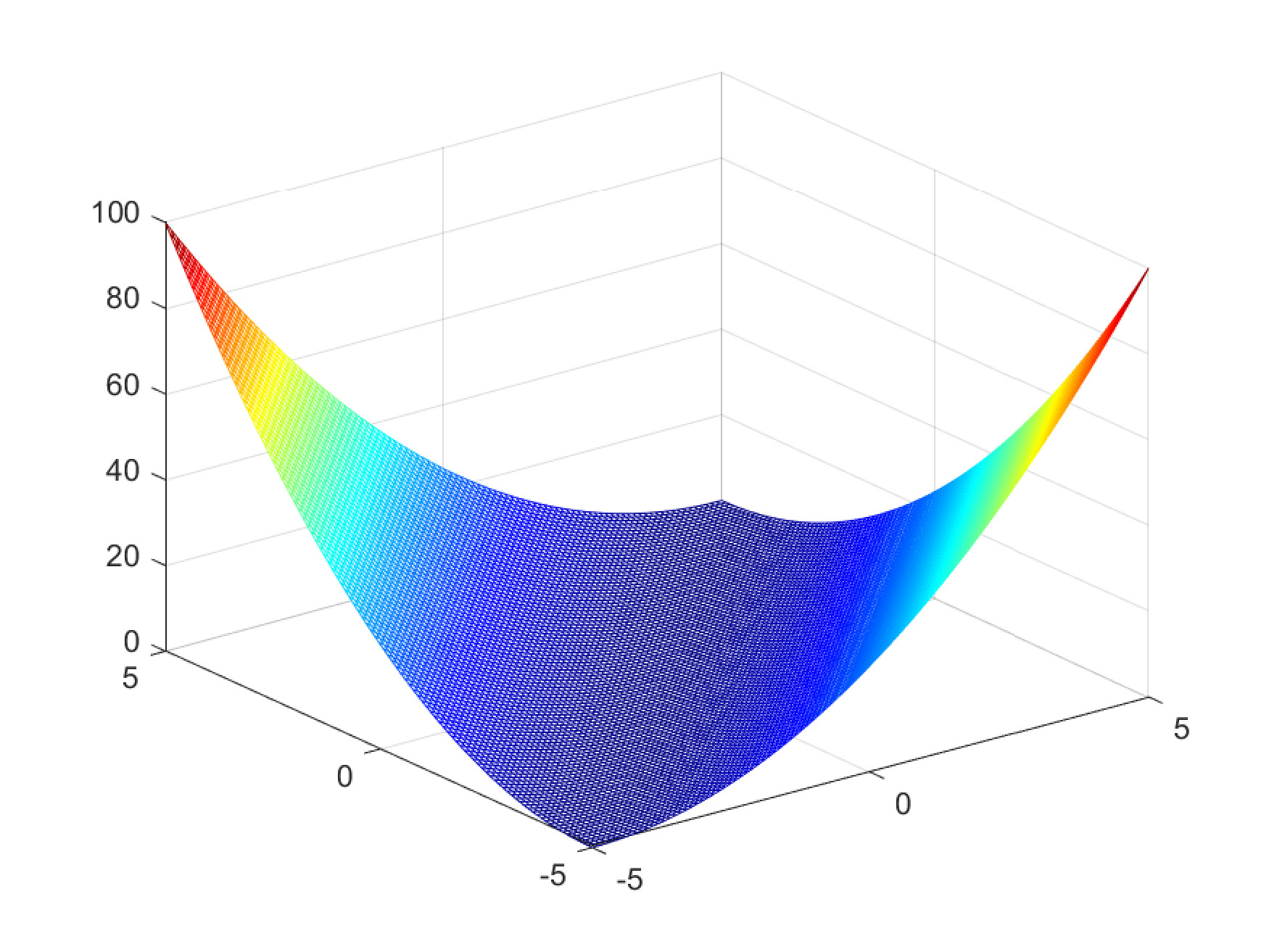} \label{MSE}}
\hspace{0.05in}
\subfigure[Correntropy]{\includegraphics[height=3cm]{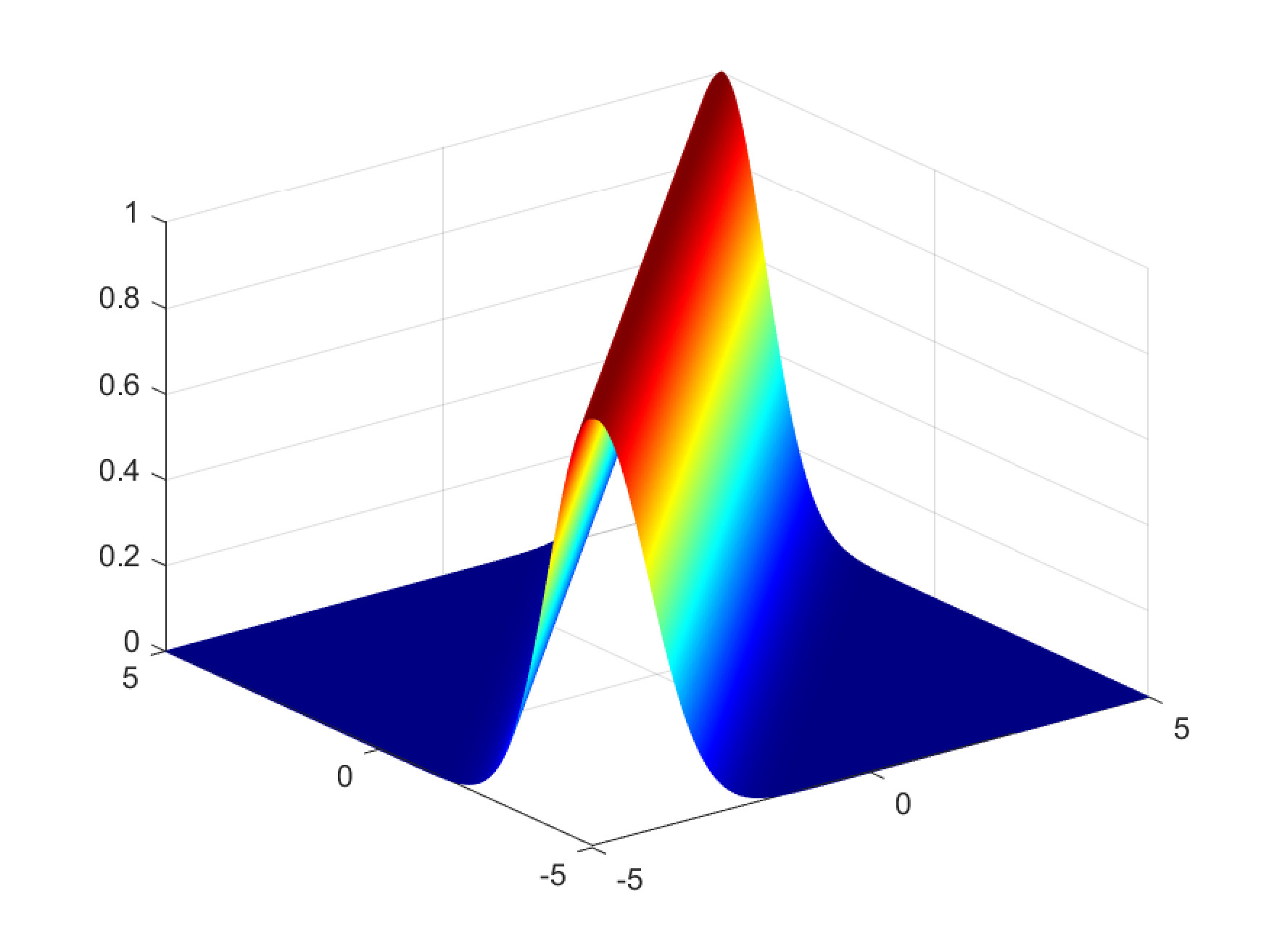} \label{Corr}}
\caption{ Illustration of MSE and Correntropy. Compared to MSE, correntropy is not sensible to outliers as it drops steadily while being far away from the central line.}
\vspace{-1em}
\label{Fig:MSE&Corr}
\end{figure}

In most variants of ICP, mean squared error (MSE) is used to measure the the residual error. These methods cannot efficiently resist the presence of heavy outlier. Fig.~\ref{MSE} illustrates an MSE-kind error function in the joint space. MSE is a quadratic function, it takes small values in the bottom of the valley and equals 0 on the $x=y$ line. But with the difference between $x$ and $y$ getting bigger, the MSE function rises quadratically and have an amplification effect on the residual error. This makes MSE capable of dealing with distributions like Gaussian but not proper for heavy-tail distributions. The latter is exactly the case when outliers present in points clouds\cite{6816032}.

An illustration of correntropy is shown in Fig.~\ref{Corr} in the joint space. As a matter of fact, correntropy emphasizes the error close to the $x=y$ line, and drops steadily while being far away from it, which means correntropy can restrict the residual error from been excessively amplified and is more suitable for being a similarity measure for non-Gaussian variables.

As stated by Hasanbelliu et al. in \cite{6816032}, with a finite number of points $\left\{{\bf x}_{i},{\bf y}_{i}\right\}_{i=1:N}$, a sample estimator of correntropy can be drawn as:
\begin{equation}
    \hat{v}\left(X,Y\right)=\frac{1}{N}\sum\limits_{i=1}^{N}
    G_{\sigma}\left({\left\|{\bf x}_{i}-{\bf y}_{i}\right\|_2^2}\right),
    \label{Eq Informatic Corr}
\end{equation}
where $G_{\sigma}\left(r^2\right)=\frac{1}{\sqrt{2\pi}\sigma}\exp{\left(-\frac{r^2}{2\sigma^2}\right)}$ is the density function of the Gaussian kernel with the bandwidth parameter $\sigma$. For more information about the derivation, please refer to \cite{6816032}.

Correntropy is known to be robust to non-Gaussian noise and it has been introduced to the ICP algorithm \cite{du2018robust}. Though they are widely used in robotics, the concept of entropy and parameter tuning are often confusing and can be challenging in real world applications. However, we discover that correntropy is equivalent to noise modeling in NDT. Therefore, it is possible to apply knowledge relating to NDT to correntropy metrics directly. 

\begin{figure}
\centering
\subfigure[Likelihood]{\includegraphics[height=3cm]{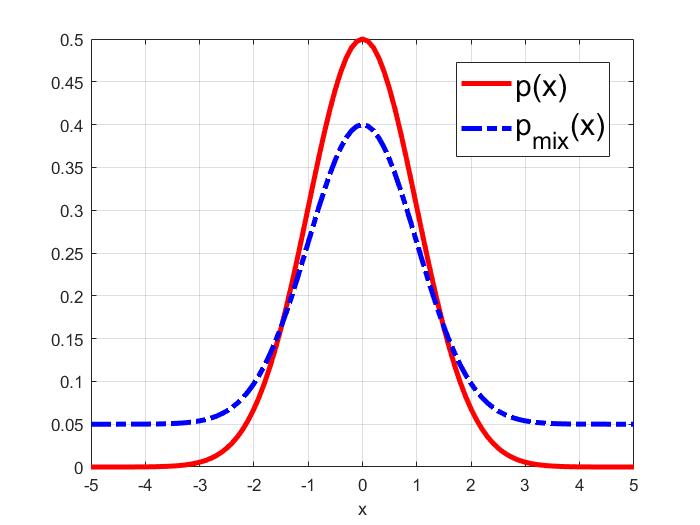} \label{Likelihood}}
\hspace{0.05in}
\subfigure[Negative log-likelihood]{\includegraphics[height=3cm]{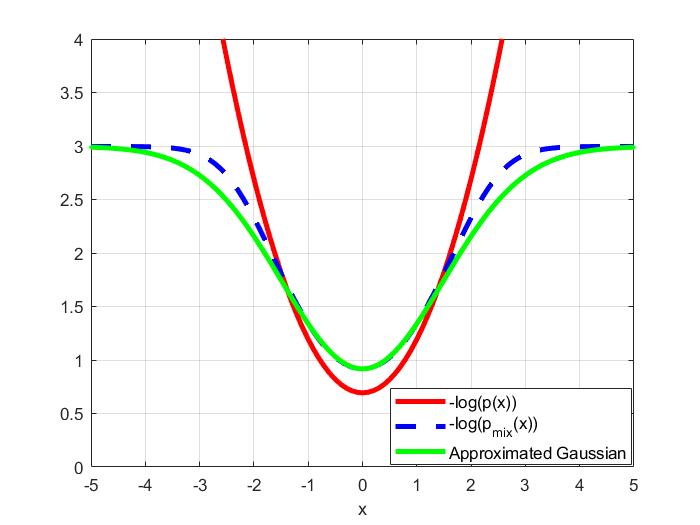}
\label{Negative log-likelihood}}
\caption{ Illustration of Likelihood and Negative log-likelihood. The green line shows the effectiveness of the Gaussian approximation to $p_{mix}(x)$.}
\label{Fig:Likelihood}
\end{figure}

If we consider a pair of points $\left({\bf a}_{i},{\bf b}_{i}\right)$ is associated by bidirectional search $\mathcal{I}_{bi}$. The relationship between the two points can be drawn as ${\bf a}_{i}=\mathbf{T^{\ast}}{\bf b}_{i}+{\bm \epsilon}_i$, where ${\bm \epsilon}_i$ is the sensor error. It is well-handled by the noise modeling of GICP. While the point-pair belongs to inlier, the Gaussian assumption of the pair-wise residual error is valid. However, when one of them belongs to outlier, the pair-wise residual error ${\bf e}_{i}={\bf a}_{i}-\mathbf{T^{\ast}}{\bf b}_{i}-{\bf \epsilon}_{i}$ cannot be properly modeled by a single Gaussian distribution, but a mixture of a Gaussian and a uniform distribution:
\begin{equation}
    {\bf e}_{i}\sim\mathcal{N}\left({\bf 0},\bf{\Omega}_{i}\right)+\mathcal{U}_{i},
\end{equation}
or equivalently as:
\begin{equation}
    p_{mix}\left({\bf e}_{i}\right)=c_1\exp{\left(-\frac{{\bf e}_{i}^\top\bf{\Omega}_{i}{\bf e}_{i}}{2}\right)}
    +c_2p_0,
    \label{Eq:G+uni}
\end{equation}
where $p_0$ is the ratio of outlier. Using MLE, the objective function can be derived as:
\begin{equation}
    \mathbf{T}^{\ast}\!=\!\mathop{\arg\max}_{\mathbf{T}}\prod\limits_{i=1}^{N}
    p_{mix}\left({\bf e}_{i}\right)\!=\!
    \mathop{\arg\min}_{\bf T}\sum\limits_{i=1}^{N}-\log{p_{mix}\left({\bf e}_{i}\right)}
    \label{Eq Negative log-likelihood}
\end{equation}
The PDF in (\ref{Eq:G+uni}) do not have a good mathematical form in the optimization perspective as the formed objective function has no simple first- and second-derivatives therefore we use the approximation provided by \cite{1249285} to the negative log-likelihood in (\ref{Eq Negative log-likelihood}): 
\begin{equation}
    -\log{\left(p_{mix}\left({\bf e}_{i}\right)\right)} \approx d_1\exp\left(-d_2\frac{{\bf e}_{i}^\top\bf{\Omega}_{i}{\bf e}_{i}}{2}\right)+d_3,
    \label{Eq appro}
\end{equation}
where $d_1$, $d_2$ and $d_3$ are computed by requiring the approximation function behaves similarly to $-\log{p_{mix}\left({\bf e}_{i}\right)}$ at extreme values.
Fig.~\ref{Negative log-likelihood} shows shows the validity of the aforementioned approximation (the green line).

The parameter $d_3$ can be omitted as it is just an offset and has no affect on solving the minimum of a summation objective function in (\ref{Eq Negative log-likelihood}). 

Comparing (\ref{Eq appro}) to (\ref{Eq Informatic Corr}), they have a same kind of mathematical formulation with a scaled square of the Euclidean distance inside of the {\it exp} term. The bandwidth parameter $\sigma$ in (\ref{Eq Informatic Corr}) plays the same role as $d_2$ in (\ref{Eq appro}). For NDT, the outlier ratio is fixed, which is equivalent to a fixed $\sigma$ in correntropy. So we draw the conclusion that correntropy is equivalent to noise modeling in NDT. Thanks to the insight in optimization in correntropy based applications, the $\sigma$ should be gradually changed during the optimization, which is equivalent to adaptively tuning the outlier ratio for NDT. Therefore, the proposed CoBigICP has the potential to be more robust and accurate than NDT.


To recap briefly, the pair-wise error can be view as Gaussian if they are inliers and cannot if they are outliers. Since we can never have the perfect correspondence, the error distribution may not be Gaussian and MSE is not the optimal. We use a more delicate noise model that is a mixture of a Gaussian distribution and a uniform distribution similar to NDT. As proved before, this is actually a correntropy-based noise modeling technique. So, with the bidirectional search $\mathcal{I}_{bi}$ the objective function of the proposed method is formulated as:
\begin{align}
    \begin{split}
    \mathbf{T}^{\ast} &=\mathop{\arg\max}_{\mathbf{T}}\sum\limits_{i=1}^{N}
    G_{\sigma}(r_{i}), \\
    r_{i}^2 & ={\left({\bf a}_{i}-\mathbf{T}{\bf b}_{i}\right)^\top\bf{\Omega}_{i}\left({\bf a}_{i}-\mathbf{T}{\bf b}_{i}\right)}.
    \end{split}
    \label{Eq ProObFun}
\end{align}
Note that the sign of the regularization term is changed due to the different sign of MSE and correntropy. The bandwidth parameter $\sigma$ can be tuned easily to model the portion of outlier. As shown in Hasanbelliu's work \cite{6816032}, one feasible approach would be to change the kernel size with a decay rate during optimization. To this end, our approach is more flexible than a fixed portion of outlier.

\subsection{Lie-Algebra-Based Solver}
The point set registration problem is finally converted to optimize a smooth function on a differentiable manifold. To solve the optimization problem in (\ref{Eq ProObFun}), traditional approaches are based on gradient-descent and these methods usually have slow rate of convergence. We propose a lie-algebra-based solver that can convert the problem into an iteratively reweighted least squares (IRLS) and the solution can be found in analytical form. The proposed solving process is based the standard workflow of Riemannian based optimization method in \cite{absil2007trust}. The solution $\mathbf{T}$ is actually on a Riemannian manifold: special Euclidean group $SE(3)$ and the rotation matrix $\bf{R}$ belongs to special orthogonal group SO(3), which can be defined as follows:
$SO(3)\overset{\text{def}}{=}\left\{{\bf R}\in \mathbb{R}^{3\times3}:{\bf R}^\top{\bf R}={\bf{I}},det({\bf R})=1\right\}$, $SE(3)\overset{\text{def}}{=}\left\{\mathbf{T}\in\begin{pmatrix} {\bf R} & {\bf t} \\ 0 & 1 \end{pmatrix}:{\bf R}\in SO(3), {\bf t}\in \mathbb{R}^3\right\}$. The group $SO(3)$ forms a smooth manifold. The tangent space of the manifold is denoted as $\bf{so}(3)\in\mathbb{R}^3$, which is also called the Lie Algebra ($\bf{se}(3)\in\mathbb{R}^6$ as well). $\bf{so}(3)$ also coincides with $3\times3$ skew symmetric matrices, denoted by the {\it hat} operator:
\begin{align}
    {\bf \phi}^{\wedge}=\begin{bmatrix} \phi_1 \\ \phi_2 \\ \phi_3 \end{bmatrix}^{\wedge}
    =\begin{bmatrix}0 & -\phi_3 & \phi_2 \\ \phi_3 & 0 & -\phi_1 \\ -\phi_2 & \phi_1 & 0\end{bmatrix}\in so(3)
\end{align}

A standard approach for optimization problem on manifold \cite{absil2007trust} consists of determining a retraction $\mathcal{R}_{\mathbf{T}}$. The above function is reparametrized as:
\begin{equation}
    \mathop{\max}_{\mathbf{T}\in SE(3)}f({\bf T})\Rightarrow
    \mathop{\max}_{\delta {\bf x}\in \mathbb{R}^n}f(\mathcal{R}_{\mathbf{T}}(\delta {\bf x})).
    \label{Eq retraction}
\end{equation}

As stated by Forster in \cite{10.1109/TRO.2016.2597321}, the use of the retraction allows framing the optimization problem over an Euclidean space of suitable dimension, e.g.  $\delta {\bf x}\in \mathbb{R}^3$ when we work in $SO(3)$ and $\delta {\bf x}\in \mathbb{R}^6$ when we work in $SE(3)$. Retraction is actually a bijective map between a perturbation $\delta {\bf x}$ of the tangent space at $\bf x$ and a neighborhood of $\bf x$ on the manifold. And after all, the right hand side of (\ref{Eq retraction}) can be formulated as an IRLS. By solving the IRLS, we get a vector $\delta{\bf x}^{\ast}$ in the tangent space. Then the current guess on the manifold, which is $\bf x$, can be updated.

The {\it exponential map} associates the lie algebra to a rotation has a first-order approximation: $exp({\bm \phi}^{\wedge})\approx \bf{I}+{\bm \phi}^{\wedge}$. {\it Exponential map} can be a retraction. It is not the computational optimal according to \cite{10.1109/TRO.2016.2597321}. In this paper, we use the following retraction for $SE(3)$:
\begin{equation} 
    \mathcal{R}_{\mathbf{T}}(\delta {\bf x})=\left(exp( {\bf \xi}^{\wedge}){\bf R},exp( {\bf \xi}^{\wedge}){\bf t}+\Delta {\bf t}\right), 
    \label{Eq SE(3) retraction}
\end{equation}
where $\delta{\bf x}=\begin{bmatrix}{\bf \xi} \\ \Delta {\bf t}\end{bmatrix}\in \mathbb{R}^6$. 
$\mathcal{R}_{\bf T}(\delta {\bm \xi},\delta {\bf t})$ means we use retraction at ${\bf T}\doteq({\bf R}, {\bf t})$.

By applying the first-order approximation: $exp({\bf \phi}^{\wedge})\approx \bf{I}+{\bf \phi}^{\wedge}$ to (\ref{Eq SE(3) retraction}), we get the perturbed version of $\bf R$ and $\bf t$:
\begin{equation}
    \mathcal{R}_{\bf T}(\delta {\bf x})=\left(\left({\bf{I}}+{\bf \xi}^{\wedge}\right){\bf R},\left({\bf{I}}+{\bf \xi}^{\wedge}\right){\bf t}+\Delta {\bf t}\right).
    \label{Eq sf retracrion}
\end{equation}

We can now reparametrize the pair-wise error in (\ref{Eq ProObFun}):
\begin{align}
    \begin{split}
    {\bf a}_{i}-{\bf R}{\bf b}_{i}-{\bf t} & = {\bf a}_{i}-\left({\bf{I}}+{\bf\xi}^{\wedge}\right){\bf R}{\bf b}_{i}-\left({\bf{I}}+{\bf \xi}^{\wedge}\right){\bf t}-\Delta {\bf t} \\
    & = {\bf v}_{i}+{\bf H}_{i}\delta{\bf x},
    \end{split}
    \label{lie pair}
\end{align}
with $\bf{{\bf v}}_{i}={\bf a}_{i}-{\bf Rb}_{i} - {\bf t},
    {\bf H}_{i}=\begin{bmatrix}({\bf Rb}_{i}+{\bf t})^{\wedge} -{\bf{I}}\end{bmatrix}$.
In the derivation of (\ref{lie pair}), the property of cross product is used: ${\bf a}^{\wedge}{\bf b}=-{\bf b}^{\wedge}{\bf a}$.
Now that (\ref{Eq ProObFun}) has the following form:
\begin{align}
    \begin{split}
    \delta{\bf x}^{\ast}&=\mathop{\arg\max}_{{\bf x}}\sum\limits_{i=1}^{N}
    w_{G}{\left({\bf v}_{i}+{\bf H}_{i}\bf{x}\right)^{\mathrm{T}}\bf{\Omega}_{i}\left({\bf v}_{i}+{\bf H}_{i}\bf{x}\right)} \\
    &=\mathop{\arg\max}_{{\bf x}}\left({\bf x}^{\mathrm{T}}\mathbf{A}{\bf x}+2{\bf b}^{\mathrm{T}}{\bf x}+const\right) \\
    &=-\mathbf{A}^{\dagger}{\bf b},
    \end{split}
\end{align}
where $\mathbf{A}=\sum\limits_{i=1}^{N}w_G{\bf H}_{i}^{\mathrm{T}}\bf{\Omega}_{i}{\bf H}_{i}$,
${\bf b}=\sum\limits_{i=1}^{N}w_G{\bf H}_{i}^{\mathrm{T}}\bf{\Omega}_{i}{\bf v}_{i}$ and $w_{G}=G_\sigma\left(r_i\right)$ is the weight function (\ref{Eq Informatic Corr}). 

Importantly, $\mathbf{A}^{\dagger}$ denotes the pseudoinverse matrix of $\mathbf{A}$. This aims at dealing with the situation where $\mathbf{A}$ being rank-deficient.

The derivation of the above function is based on the assumption that in each iteration, the rotation matrix in the information matrix in (\ref{Eq ProObFun}) can be approximate by the result of the last iteration. With this assumption, the objective function of the Lie Algebra based optimization problem remains quadratic. Consequently, the result is in an elegant closed-form and there's no need to use gradient based methods. After we get $\delta{\bf x}^{\ast}$, the current estimation on $SE(3)$ can be updated by using the retraction (\ref{Eq SE(3) retraction}) again.

\begin{figure*}
\centering
\subfigure[ECDF of translation error] {\includegraphics[height=4.8cm]{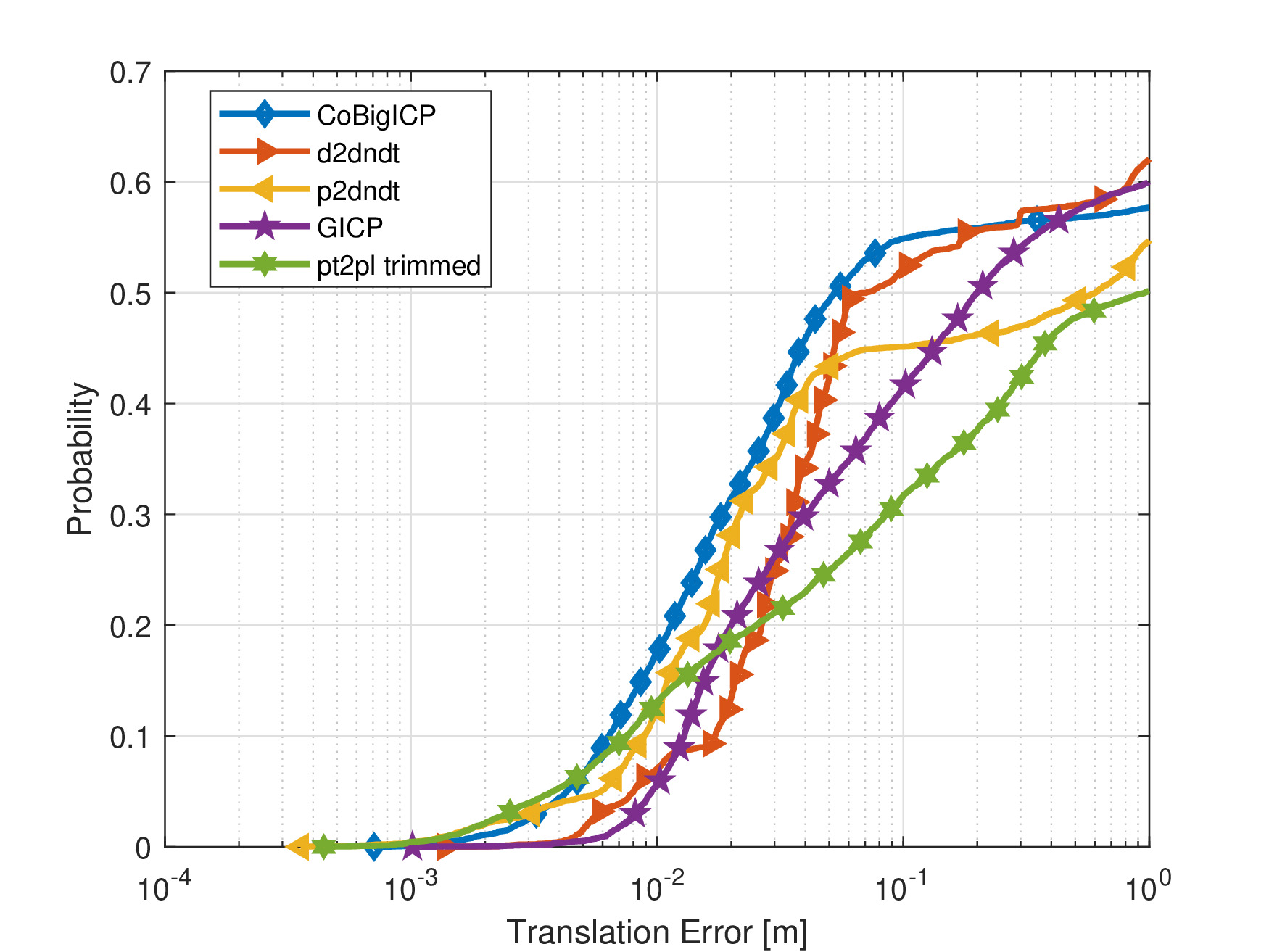}}
\hspace{0.6in}
\subfigure[ECDF of rotation error] {\includegraphics[height=4.8cm]{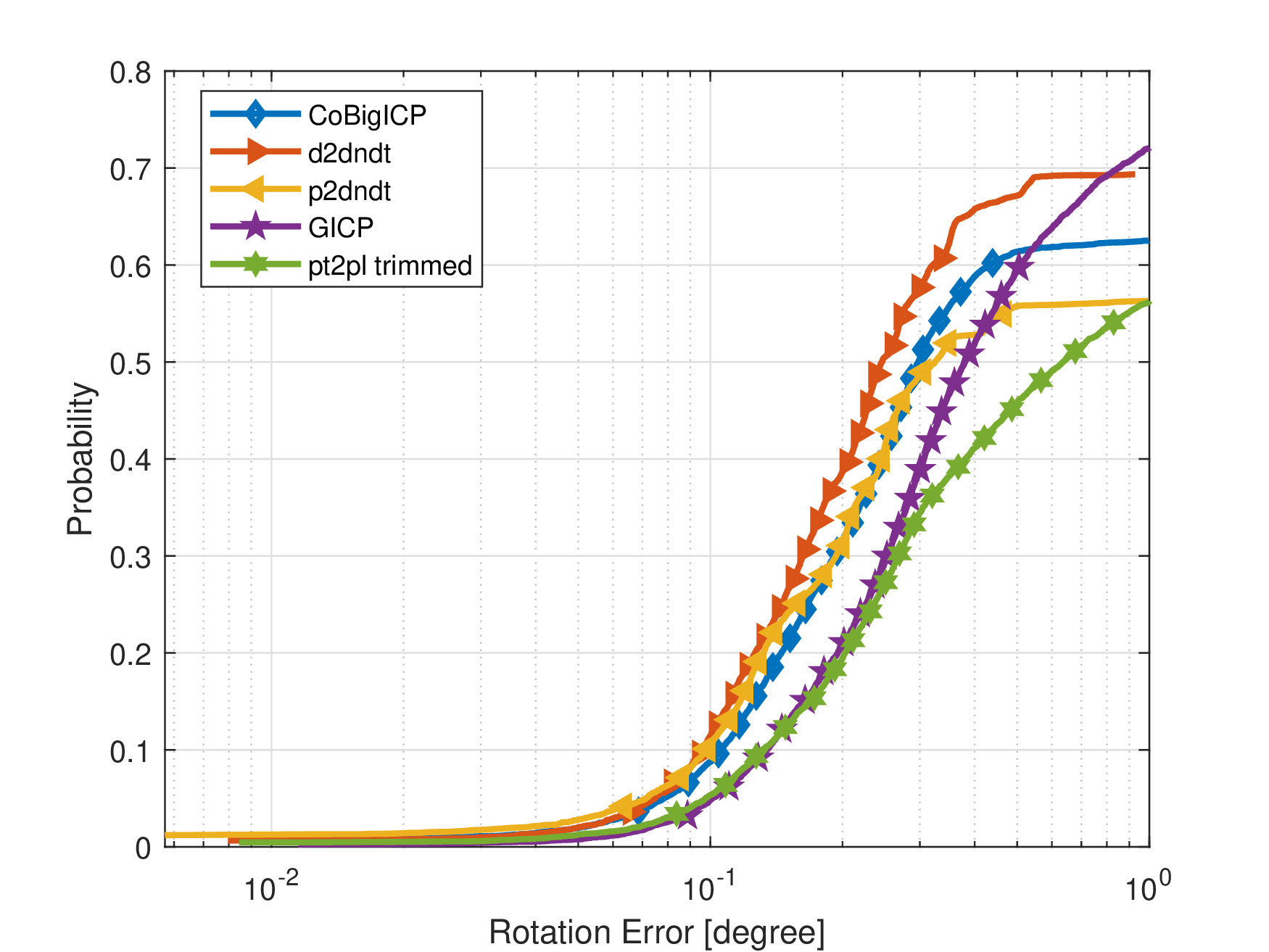}}
\caption{ The empirical cumulative distribution function (ECDF) of different methods in ETH Hauptgebaude. The x axis represents the translation error or the rotation error while the corresponding y axis represents the probability of producing this error.}
\label{Fig ECDF}
\vspace{-1em}
\end{figure*}

\section{Experiments}
We conduct a variety of experiments to test the robustness, accuracy and efficiency of the proposed method and to compared our system with several state-of-the-art methods. A group of well-known and public available data sets are used: the ETH Challenging data sets \cite{Pomerleau:2012}.

To measure the performance of an algorithm, we use the protocol provided by \cite{10.1007/s10514-013-9327-2}. More specifically, the difference between the estimated transformation matrix $\mathbf{\hat{T}}$ and the ground truth transformation matrix $\mathbf{T}$ is computed by $\Delta\mathbf{T}=\mathbf{\hat{T}}\cdot\mathbf{T}^{-1}$. Then the translation error, also known as Relative Pose Error (RPE), and rotation error is computed by the following equations: $e_{trans}=\sqrt{\Delta{x}^2+\Delta{y}^2+\Delta{z}^2}$ and $e_{rot}=\arccos{\left(\frac{trace(\Delta{{\bf T}})}{2}-1\right)}$, where $\left(\Delta{x},\Delta{y},\Delta{z}\right)$ are the elements on the last column. 

The proposed algorithm was implemented in MATLAB (v2019b) on a PC with Intel Core i5-8300H @ 2.30GHz and 8G RAM. For fair comparison, we use the NDT implementation in MATLAB. Since GICP is closely related to our algorithm, we use our own implementation with reference to the version in PCL. We remark that GICP shares the same surface statistics (covariance matrix, information matrix and surface normal) with CoBigICP. For other involved methods, we use the implementation that the authors provide. In all experiments, 
as mentioned in the protocol of the data set, point clouds are down-sampled to 10$\%$. For the bandwidth parameter $\sigma$, we adopt result in \cite{6816032} with a decay rate of 0.97.

\subsection{Consecutive Registration and Ablation Study}
In this section we conduct experiments of all methods on a series of consecutive pairs of scans in two data sets to evaluate their performances under both structured and semi-structured environment: 
\begin{itemize}
\item[-] $\bm{ETH Hauptgebaude}$ collects lidar frames along a corridor with repetitive pillars and arches. 
\item[-] $\bm{Gazebo Summer}$ is a semi-structured environment, with a mixture of man-made structures, vegetation and some dynamic objects like the walking people.
\end{itemize}
We compared CoBigICP with GICP, NDT and MiNoM \cite{8594278} by computing the RPE for each consecutive pair. Main results are shown in table \ref{tabETH} and \ref{tabGazebo}. Note that the method {\it CoGICP} aims at ablation study which discards the bidirectional correspondence and involves only the correntropy metric. Compared to other methods, CoBigICP is the most competitive with the best accuracy and favorable time consumption. In the ETH Hauptgebaude data set, CoBigICP far more exceeds the other algorithms and this phenomenon supports the importance of correspondence step while also validating our bidirectional correspondence that we propose in this paper. The result of Gazebo Summer shows CoBigICP is also capable for semi-structured environment. The ablation study between CoGICP and CoBigICP shows effectiveness of bidirectional correspondence in both accuracy and speed. An illustrative reconstruction result is shown in Fig.~\ref{Fig recon res}. We compute the pose estimation using CoBigICP and we put the transformed point cloud together. The reconstruction result (colored points) is consistent with the ground truth (white points).

\subsection{Convergence basin under perturbations}
In this section, 6720 scan pairs from ETH Hauptgebaude (not necessarily consecutive) with different initial pose perturbations provided by Pomerleau et al. \cite{10.1007/s10514-013-9327-2} are used to evaluate the robustness or convergence basin of algorithms. The different hardness of perturbations are indicated by {\it easyPose, midiumPose} and {\it hardPose}. These perturbations makes the registration challenging and meaningful. To lead a fair comparison, results of d2dndt, p2dndt and trimmed-pt2pl \cite{1047997} are provided by the authors and are downloaded directly from the website. Results in Fig.~\ref{Fig ECDF} show that CoBigICP has the widest translation convergence basin while being also competitive with rotation perturbation. The phenomenon is consistent with our analysis in Section \uppercase\expandafter{\romannumeral3} {\it B}.

\begin{table}
\caption{Results for ETH Hauptgebaude}
\begin{center}
\begin{tabular}{cccccc}
\hline
\textbf{Methods} & \textbf{\textit{GICP}}& \textbf{\textit{NDT}} & \textbf{\textit{MiNoM}}& \textbf{\textit{CoGICP}}$^{\mathrm{a}}$ & \textbf{\textit{CoBigICP}} \\
\hline
$e_{trans}$[m]$\downarrow$& 0.03 & 0.16 & 0.46 & 0.02 & $\bm{0.005}$ \\
\hline
$e_{rot}$[degree]$\downarrow$& 0.19 & 9.08 & 0.32 & 0.19 & $\bm{0.15}$\\
\hline
time[s]$\downarrow$& 2.10 & 2.94 & 2.09 & 1.55 & $\bm{0.64}$ \\
\hline
\multicolumn{5}{l}{$^{\mathrm{a}}$For ablation study.}
\end{tabular}
\end{center}
\label{tabETH}
\vspace{-1em}
\end{table}

\begin{table}
\caption{Results for Gazebo Summer}
\begin{center}
\begin{tabular}{cccccc}
\hline
\textbf{Methods} & \textbf{\textit{GICP}}& \textbf{\textit{NDT}} & \textbf{\textit{MiNoM}}& \textbf{\textit{CoGICP}}$^{\mathrm{a}}$ & \textbf{\textit{CoBigICP}} \\
\hline
$e_{trans}$[m]$\downarrow$& 0.10 & 0.16 & 0.36 & $\bm{0.05}$ & 0.06 \\
\hline
$e_{rot}$[degree]$\downarrow$& 1.76 & 9.86 & 3.19 & 0.34 & $\bm{0.21}$\\
\hline
time[s]$\downarrow$& 2.43 & 2.32 & 2.00 & 1.08 & $\bm{0.87}$ \\
\hline
\multicolumn{5}{l}{$^{\mathrm{a}}$For ablation study.}
\end{tabular}
\end{center}
\label{tabGazebo}
\vspace{-1em}
\end{table}

\section{Conclusions}
In this paper, we present a novel point cloud registration algorithm based on bidirectional correspondence and correntropy. Extensive comparative experiments showing our method reaches state-of-the-art precision and offers better robustness even with poor initial values.

In the future, we plan to study the adaptive estimation of outlier ratio by the robot's travelling distance between consecutive frames and predictable number of dynamic objects in the robot's working environments. 
And finally remove the trade off between parameter selection and the performance.

\bibliographystyle{IEEEtran}
\bibliography{mybibfile}

\end{document}